\pdfoutput=1

\documentclass[11pt]{article}

\usepackage[final]{acl}

\usepackage{times}
\usepackage{latexsym}
\usepackage{listings}
\usepackage{xcolor}
\usepackage{tcolorbox}
\tcbuselibrary{listings}
\usepackage{tcolorbox}
\usepackage{tabularx}
\usepackage{ragged2e}
\newcolumntype{Y}{>{\RaggedRight\arraybackslash}X}
\tcbuselibrary{listings,breakable}
\usepackage[T1]{fontenc}

\usepackage[utf8]{inputenc}

\usepackage{microtype}

\usepackage{inconsolata}

\usepackage{graphicx}

\newcommand*{\affmark}[1][*]{\textsuperscript{#1}}

\renewcommand{\lstlistingname}{Prompt}

\title{\textit{ScheduleMe}: Multi-Agent Calendar Assistant}

\author{Oshadha Wijerathne\affmark[1], Amandi Nimasha\affmark[1], Dushan Fernando\affmark[1],\\ \textbf{Nisansa de Silva\affmark[1], \and Srinath Perera\affmark[2]} \\
  \affmark[1]Department of Computer Science \& Engineering, 
  University of Moratuwa, 
  Sri Lanka \\
  \texttt{\{oshadha.20, amandi.20, dushan.20, NisansaDdS\}@cse.mrt.ac.lk}\\
   \affmark[2]WSO2 LLC\\
  \texttt{srinath@wso2.com} 
  }

\begin{document}
\maketitle
\begin{abstract}
Recent advancements in LLMs have contributed to the rise of advanced conversational assistants that can assist with user needs through natural language conversation. This paper presents a \textit{ScheduleMe}, a multi-agent calendar assistant for users to manage google calendar events in natural language. The system uses a graph-structured coordination mechanism where a central supervisory agent supervises specialized task agents, allowing modularity, conflicts resolution, and context-aware interactions to resolve ambiguities and evaluate user commands. This approach sets an example of how structured reasoning and agent cooperation might convince operators to increase the usability and flexibility of personal calendar assistant tools.  
\end{abstract}

\section{Introduction}

The rapid advancements in natural language processing (NLP) and large language models (LLMs) have opened new opportunities for developing intelligent, user-friendly applications. Among these, conversational agents capable of understanding and acting upon human language inputs are becoming increasingly important for daily task management. Traditional scheduling systems often require rigid, form-based inputs and manual navigation, limiting user experience and efficiency. In contrast, leveraging LLMs enables the creation of systems that can interpret flexible, natural language instructions, offering a more intuitive and seamless interaction.

This research introduces a calendar management assistant built using the LangChain framework \citep{folstad2019chatbots} and OpenAI’s GPT-4o mini model \citep{openai2024gpt4technicalreport,wang-2025-optimizing}.The assistant enables users to manage their calendars through natural, conversational interactions.

The system adopts a modular, multi-agent architecture. Specialized agents handle specific operations such as scheduling, fetching, editing, and deleting events. A centralized supervisory chatbot coordinates these agents and manages the dialogue with the user. This separation improves modularity, reliability, and context-aware task execution.

The goal is to create an assistant that not only executes tasks accurately based on natural user requests, but also enhances transparency, reliability, and user satisfaction. Users will be able to manage their calendars simply by conversing with the chatbot, eliminating the need for complex interfaces or strict command formats. Through this research, our objective is to demonstrate how LLMs, when properly structured within a robust framework such as LangChain, can serve as powerful tools for building practical, real-world intelligent applications that align with human communication patterns.

\section{Related Work}
\subsection{Limitations of Traditional Scheduling Dialogue Systems}

Traditional task-oriented dialogue systems typically rely on intent classification and slot-filling methods \citep{Surdeanu2011Stanford}. These systems map user inputs to predefined actions and extract key details such as date, time, or participants.

Dialogue flows are often rigid and frame-based \citep{braggaar2024evaluatingtaskorienteddialoguesystems}, collecting user input step-by-step. While effective in simple cases, this structure struggles when users provide information out of order, revise commands, or use unexpected phrasing.
Systems such as \textit{Calendar.help} \citep{cranshaw2017calendar} have shown how these methods can be used in real-world scheduling applications, combining natural language understanding with backend tools such as webhook integrations. However, these systems often fail to handle ambiguous or incomplete inputs gracefully and are difficult to adapt to new use cases without retraining on labeled data. As a result, their user interactions can feel rigid and frustrating.

\subsection{Advancing to LLM-Based Multi-Agent Systems}

Earlier AI agents were designed using symbolic rules or simple learning methods were built to act independently, make decisions, and sometimes communicate with other agents~\citep{farinetti2024chatbot,alonso2002ai,hazra2024saycanpayheuristicplanninglarge}. Although these early agents were effective in narrow tasks, they required a lot of manual programming and were not very flexible~\citep{Wang_2024}.

Recent advances in large language models (LLMs), such as GPT-4  \citep{openai2024gpt4technicalreport,qin2023chatgptgeneralpurposenaturallanguage}, PaLM  \citep{JMLR:v24:22-1144}, and LLaMA \citep{touvron2023llamaopenefficientfoundation}, have introduced new possibilities. These models can handle open-ended language tasks, reason about goals, and even use external tools when guided properly. Methods such as chain-of-thought prompting~\citep{NEURIPS2022_9d560961}, in-context learning, and tool use through APIs have enabled LLM-based agents to solve more complex and varied problems~\cite{warnakulasuriya2025evolution}.

Frameworks such as ReAct \citep{yao2023reactsynergizingreasoningacting} and AutoGPT \citep{yang2023autogptonlinedecisionmaking} have demonstrated how LLMs can be used as the 'brains' behind autonomous agents.However, many tasks such as scheduling or workflow management require multiple specialized agents to work together. A multi-agent setup allows for modular design, parallel task execution, and clearer delegation of responsibilities. Without structured coordination, these systems often struggle with effective communication, shared memory, and maintaining context, which limits their ability to handle complex or extended interactions.

\subsection{Multi-Agent Approaches to Scheduling Assistants}

Some recent systems split scheduling tasks into smaller parts, assigning different agents to handle event creation, editing, retrieval, and similar operations. \textit{SmartCal} \citep{shen2024smartcalapproachselfawaretooluse} improves tool use reliability and decision making through a self-aware supervisory framework, which coordinates agent actions, supports error recovery, and maintains context. Without such supervision, assistants often struggle with user corrections and overlapping goals during complex interactions.

Beyond orchestration, procedural and multi-step reasoning is crucial for advanced scheduling. ScriptWorld  \citep{ijcai2023p566}  demonstrates how agents can learn and execute sequential tasks while maintaining state. Although it operates in simulation, the same principle applies to real-world scheduling, where agents must interpret goals, resolve conflicts, and complete tasks in the correct order.

To overcome the limits of static coordination, frameworks like LangGraph \citep{duan2024explorationllmmultiagentapplication, wang2024agentailanggraphmodular} provide graph-based, state-aware workflows. Nodes represent task-specific agents or tools, and edges define transitions based on system state, enabling conditional branching, iteration, and runtime adaptation. Integrating LangGraph with a central supervisory agent supports dynamic task execution, context maintenance, and natural conversational scheduling.



\section{Methodology}

\subsection{Multi-Agent System Architecture}

\begin{figure*}[htb]  
    \centering
    \includegraphics[width=\textwidth]{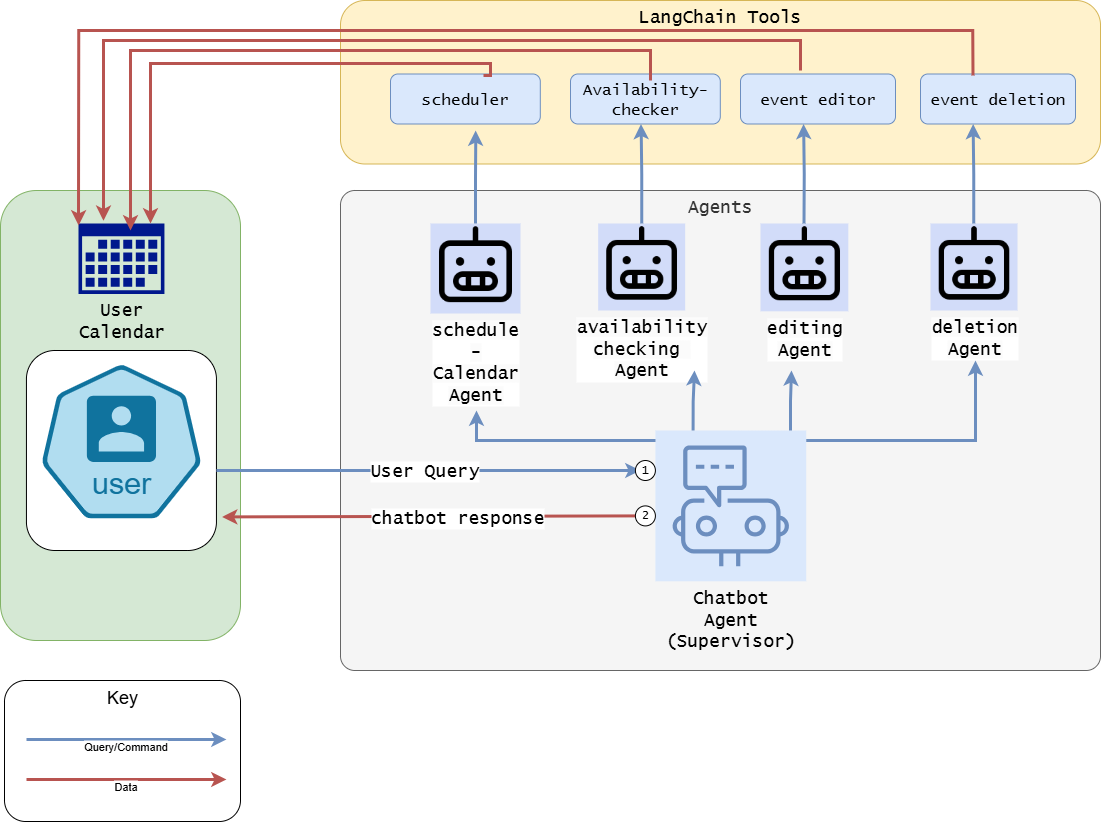}
    \caption{Multi-Agent System Architecture: All other agents are controlled by the \textit{supervisor agent}, but we have opted not to draw the control and communication lines between the agents to reduce unnecessary clutter. When a command or a data item is relevant to all the entities in a parent entity, the relevant arrow terminates on the parent entity. Otherwise, it terminates on the specific relevant child entity. The numbers on the agents at arrow terminals indicate the order in which each action may happen in a typical execution.}
    \label{fig:Multi-Agent System Architecture}
\end{figure*}

Having discussed prior work and current limitations in existing systems, we now present the architecture and design principles behind our proposed calendar assistant. Our system integrates large language model (LLM) based reasoning with a graph-driven orchestration framework using LangGraph, enabling dynamic coordination among agents.

The architecture is centered around a supervisory chatbot agent, which serves as the sole interface for user interaction and agent coordination. Upon receiving a user query, the supervisor agent interprets user intent and delegates tasks accordingly to one of the specialized functional agents. These include the scheduling agent, availability checking agent, event editing agent, and event deletion agent. All inter-agent communication is mediated through the supervisor agent. When a functional agent requires additional information to complete a task, it requests the chatbot to re-engage with the user to obtain the missing input.

Each agent follows the ReAct (Reasoning and Acting) paradigm, combining decision-making with the ability to invoke predefined tools. These tools are implemented as custom functions that interface with the Google Calendar API to perform specific actions. For instance, the availability checking tool queries calendar data for events within a given time range, while the scheduling tool creates new calendar events based on parameters such as title and  time. Similarly, the editing and deletion tools update or remove events based on event ids and user-specified criteria. These tools abstract the underlying API calls, allowing agents to focus on high-level decision logic.

LangGraph is employed to structure the agent coordination process as a directed graph. In this configuration, each node represents an agent, and edges define the flow of execution based on the outcome of reasoning steps or user input. The graph structure enforces that communication paths flow through the supervisor agent, ensuring a controlled and interpretable interaction model. This setup allows the system to flexibly handle user queries in a stateful and modular manner.
An overview of the system’s architecture is depicted in Fig.~\ref{fig:Multi-Agent System Architecture}, highlighting the agents and their interactions within the LangGraph execution framework.

\subsection{Implementation Details}

With the system architecture defined, we now describe how the AI Calendar Assistant is implemented in practice, detailing the technologies and components involved. The AI Calendar Assistant is implemented as a graph-based multi-agent system, where each node corresponds to an agent and each edge represents a possible transition in the task flow. The architecture is constructed using LangGraph’s \texttt{StateGraph} module, which supports dynamic, stateful execution paths. The central supervisory chatbot agent initializes each interaction by processing user input and extracting intent, parameters, and task directives. These outputs determine the subsequent traversal of the graph and the activation of the appropriate functional agent.

The system integrates OpenAI’s GPT-4o mini model via LangChain to perform natural language understanding and dialogue management. GPT-4 is known for high accuracy in complex reasoning but is resource-intensive and costly~\cite{gunathilaka2025automatic,siddiky2025optimizing}. GPT-4o provides comparable performance, with stronger multilingual and multimodal capabilities, while offering reduced latency and computational requirements \cite{siddiky2025optimizing,zhang2024latest}. Prior studies in dialogue system design have also employed GPT-4o mini as a reference model due to its extended (128k-token) context window, function-calling support, and low latency \cite{robino2025conversation}. In the medical domain, GPT-4o has demonstrated efficiency gains and near-human conversational response times, confirming its suitability for real-time applications \cite{zhang2024latest}. 
The supervisor agent leverages this model to interpret user queries, ask for clarification when needed, and generate tailored instructions for each functional agent. The initiation of the supervisor is shown in \lstlistingname~\ref{lst:prompt_supervisor}. 
When activated, functional agents handle tasks such as scheduling, checking availability, editing events, or deleting them. These agents perform stateless operations: they receive a structured input payload, carry out the task, and return the results to the supervisor. The supervisor then communicates the outcome back to the user. We show all the functional agent prompts in Appendix~\ref{app:agent-prompts}.

\begin{tcolorbox}[colback=gray!5, colframe=black!70,  breakable]
\scriptsize
\begin{lstlisting}[caption={Chat-bot Supervisor Prompt},,captionpos=b,breaklines=true, basicstyle=\ttfamily\scriptsize,label={lst:prompt_supervisor}]
"""You are the Supervisor Agent for an AI Calendar Assistant system.

Current date and time: {current_date_time}.

Your Responsibilities:
- Talk to the user to fully understand their request.
- Collect **all required information** before sending a task to any agent.
- Send tasks to the correct agent with complete and clear information.
- Collect responses from agents and decide the next action.

Agents you can use:
- calendar_checker_agent: To check calendar events.
- event_scheduler_agent: To add new events (REQUIRES: event title, date, and time).
- event_remover_agent: To delete events.(Should Provide the event Id.)
- event_modifier_agent: To modify/edit/update events.
- user: If you need more information.

Important Rules:
1. Greet the user and ask what they want to do.
2. If user request is unclear or missing information, ask follow-up questions (one at a time) until you have everything needed.
3. Only send a task to another agent once you have **all required information**.
4. Be friendly, clear, and simple. Ask **one question at a time**.
5. Always format your reply in JSON:
   - `next`: agent to call (`calendar_checker_agent`, `event_scheduler_agent`, `event_editor_agent`, `user`, or `FINISH`)
   - `messages`: Message content (talk to the user or explain to the agent what task to do).

**EXTRA REMINDERS:**
- For scheduling an event: you must collect **event title**, **date**, and **time**.
- For deleting: you must collect **event_ID**.
- For editing : you must collect **event title** and **what exactly to edit**.
- If something is unclear, always ask the user first instead of guessing.

Example JSON message when enough info is collected:
```json

  "next": "event_scheduler_agent",
  "messages": "Schedule an event titled 'Team Meeting' on 2025-05-01 at 10:00 AM."

\end{lstlisting}
\end{tcolorbox}

The backend infrastructure is developed using FastAPI, which provides RESTful endpoints for user interaction, session handling, and communication with the agent graph. A persistent JSON-based state file maintains context across multi-turn interactions, enabling coherent task handling. Secure integration with Google Calendar is achieved via OAuth 2.0 authentication, allowing the system to create, retrieve, modify, and delete calendar events with user consent.

Temporal data handling is managed using libraries such as \texttt{pytz} for timezone resolution and \texttt{dateparser} for parsing natural language dates and times. Each task agent is equipped with a custom tool that wraps the corresponding Google Calendar API call, encapsulating the logic required for scheduling, availability queries, editing, or deletion. These tools enforce structured data handling and reduce coupling between agents and external APIs.

A frontend interface for the assistant is developed using Streamlit, allowing users to interact with the calendar assistant through a simple web-based UI. The interface displays an updated calendar view on the right side of the page, providing users with a visual confirmation of their calendar events and any modifications made by the assistant.

\begin{figure*}[htb]  
    \centering
    \includegraphics[width=0.9\textwidth]{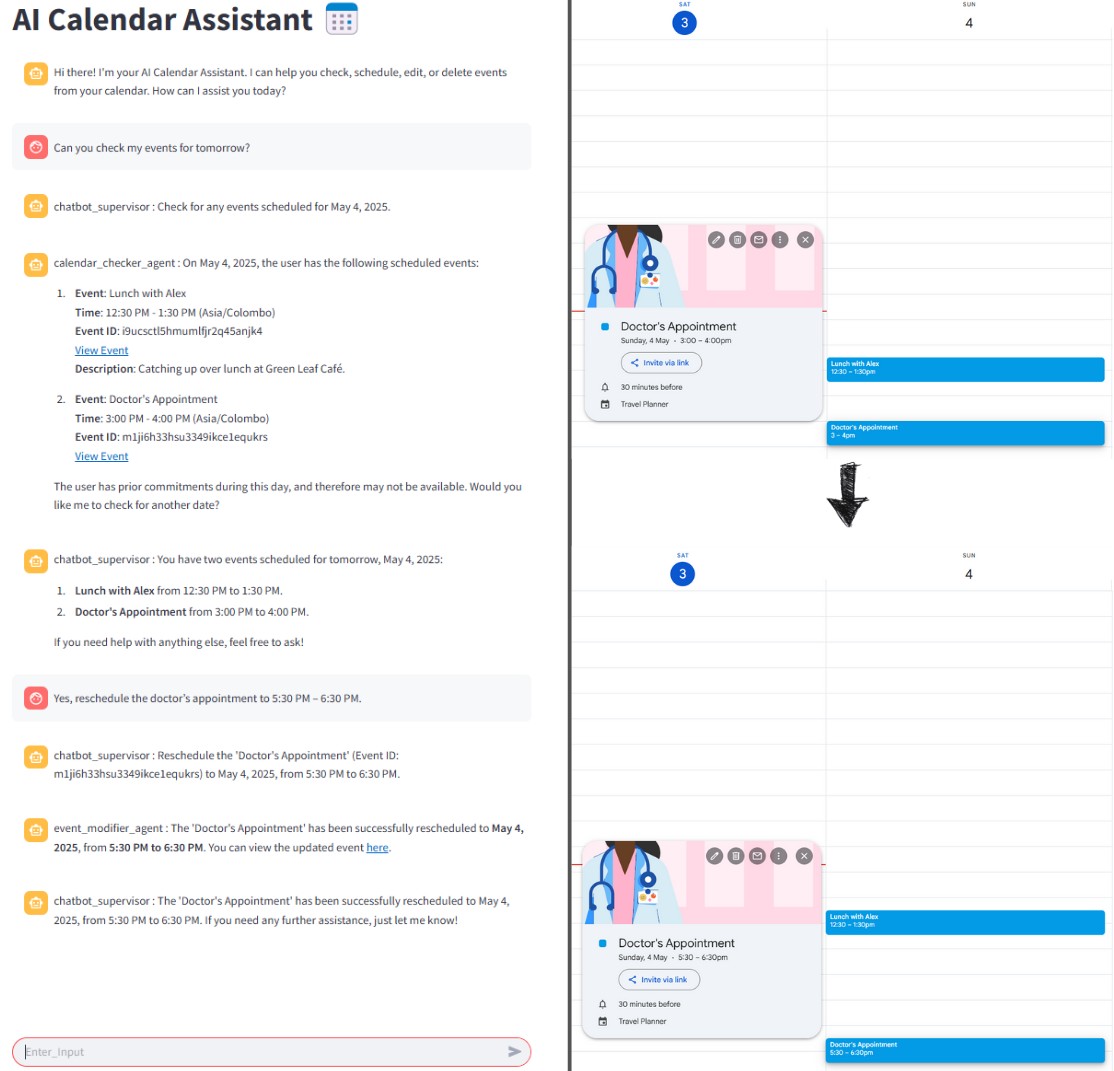}
    \caption{AI Calendar Assistant Interaction Example: A representative dialogue flow demonstrating the assistant's ability to process user queries related to calendar management. The supervisor agent interprets the user’s request and delegates actions to appropriate sub-agents (e.g., availability checking and event modification). The updated calendar view on the right confirms the successful execution of the rescheduling task.}
    \label{fig:calendar-assistant-demo}
\end{figure*}

Figure.~\ref{fig:calendar-assistant-demo} shows a representative interaction example with the AI Calendar Assistant.
The chosen stack LangGraph, LangChain, GPT-4o mini, FastAPI, Google Calendar API, and Streamlit provides a scalable, modular, and interpretable framework for enabling intelligent calendar management through natural language conversations.

\subsection{Distributed Supervisor and Scalability Enhancements}

While the core system functions well in single-user environments, real-world deployment demands scalability for concurrent users. To address this we introduce architectural extensions that support distributed execution. Specifically, we implement a distributed supervisor architecture designed for horizontal scalability and fault tolerance. In this upgraded design, multiple supervisor instances operate in parallel, each with a unique identifier and capable of independently managing user interactions. A custom load balancer orchestrates these instances by routing sessions to the least-loaded supervisor, ensuring session affinity and enabling automatic reassignment in the event of failure. This architecture eliminates the single point of failure and significantly improves throughput under high concurrency.

State sharing and coordination among supervisors are managed using Redis, which serves as a centralized store for session context and state metadata. Redis enables all supervisor instances to access and update shared session data, ensuring consistent behavior across distributed nodes. Its time-to-live (TTL) mechanism also facilitates automatic cleanup of inactive sessions, improving memory efficiency and reliability.

Each agent in the system is registered with a dynamic agent registry that allows real-time management of capabilities. This registry supports agent discovery, activation, and deactivation at runtime, and delegates tasks using a thread pool executor to avoid blocking operations. Combined with capability-based routing, this mechanism allows for flexible and scalable task delegation based on the nature of the request and the current system load.

The backend system is fully asynchronous, employing the \texttt{async/await} paradigm for non-blocking I/O. This approach enables the system to handle multiple concurrent conversations without blocking the main event loop, thereby improving response times and maximizing resource utilization. Asynchronous session handling, agent execution, and Redis-based state access together contribute to the assistant’s ability to maintain consistent user experience even under high concurrency.

Furthermore, the system is containerized using Docker Compose, supporting multiple calendar assistant instances running in parallel. These containers share the same Redis backend and are exposed via an Nginx reverse proxy that handles HTTP-level load balancing. Nginx performs round-robin request distribution, performs health checks, and enables SSL termination, ensuring both scalability and secure communication.

Lastly, we provide a metrics and monitoring endpoint that exposes real-time statistics regarding active sessions, supervisor loads, and agent utilization. This observability layer assists in system maintenance, performance tuning, and operational diagnostics in production environments.

Together, these enhancements transform the previously centralized architecture into a highly scalable, fault-tolerant, and distributed system as elaborated throughout this section, that meets the demands of real-world, multi-user environments.
\section{Experiments}

Direct quantitative benchmarking against existing scheduling systems is limited because most traditional assistants are rule-based with fixed workflows and lack publicly available evaluation datasets. ScheduleMe, being an LLM-driven multi-agent system, operates in a fundamentally different paradigm, where conventional rule-coverage or exact-match metrics are less meaningful. Therefore, we focus on a zero-shot multilingual evaluation to demonstrate practical task success while qualitatively contextualizing our system against representative scheduling approaches in prior work.

After implementing the system, we conducted a series of evaluations to assess its performance, especially in multilingual and zero-shot scenarios. The goal was to assess the model's ability to correctly interpret and execute calendar management commands in multiple languages without fine-tuning. Since the system uses OpenAI’s pretrained GPT-4o mini model, no task-specific training was performed.

Following the reasoning provided in the language comparative studies conducted by~\citet{wickramasinghe-de-silva-2023-sinhala} and by~\citet{jayatilleke2025zero}, we selected six languages for testing: English (\texttt{En}), German (\texttt{De}), French (\texttt{Fr}), Chinese (\texttt{Zn}), Tamil (\texttt{Ta}), and Sinhala (\texttt{Si}). For each language, we prepared a set of 20 test inputs, consisting of 5 examples per task type: scheduling, availability checking, editing, and deletion. This resulted in a total of 120 test cases across all languages.

Each input was a natural language command submitted via the assistant’s interface. An output was marked as correct if the assistant successfully interpreted the intent and executed the intended calendar action with the correct parameters.

To complement functional testing with real-world usability insights, we conducted a user study with 20 active digital calendar users. Participants interacted with ScheduleMe via a web-based interface linked to test Google Calendar accounts. Each participant completed 5 -7 calendar operations, including a mix of simple, complex, and multilingual requests. During the session, users recorded task success rate and error rate as objective metrics. After completing the tasks, participants completed a System Usability Scale (SUS) questionnaire and provided trust and satisfaction ratings on a five-point Likert scale. 

\section{Results}

The performance of \textbf{ScheduleMe} was evaluated using both
functional zero-shot multilingual testing and a small-scale user
study. This section presents the quantitative performance results, followed by qualitative observations and error analysis.

\subsection{Zero-Shot Multilingual Evaluation}

Table~\ref{tab:results} presents the number of correct task completions per language and the corresponding success rates. English serves as the baseline, achieving perfect accuracy across all task types. Performance is generally strong in European languages (German and French) and shows moderate degradation
in non-Latin scripts (Tamil, Sinhala, and Chinese), particularly for editing and deletion tasks. Overall, it can be observed that, other than in the case of Chinese (\texttt{Zn}), the language results aligns well with the language resource level categorization proposed by~\citet{ranathunga-de-silva-2022-languages}.

\begin{table}[htbp]
\centering
\caption{Zero-Shot Task Success Rates per Language. Each cell shows correct / total and each language has 5 inputs per task (20 total).}
\resizebox{.48\textwidth}{!}{
\label{tab:results}
\begin{tabular}{|l|c|c|c|c|c|c|}
\hline
\textbf{Language} & \textbf{Schedule} & \textbf{Availability check} & \textbf{Edit} & \textbf{Delete} & \textbf{Total} & \textbf{Success\%} \\
\hline
English (\texttt{En})  & 5/5 & 5/5 & 5/5 & 5/5 & 20/20 & \textbf{100\%} \\
French (\texttt{Fr})   & 5/5 & 5/5 & 4/5 & 4/5 & 18/20 & \textbf{90\%} \\
German (\texttt{De})   & 5/5 & 5/5 & 3/5 & 4/5 & 17/20 & \textbf{85\%} \\
Tamil (\texttt{Ta})    & 5/5 & 4/5 & 3/5 & 3/5 & 15/20 & \textbf{75\%} \\
Sinhala (\texttt{Si})  & 5/5 & 5/5 & 2/5 & 2/5 & 14/20 & \textbf{70\%} \\
Chinese (\texttt{Zn})  & 4/5 & 3/5 & 3/5 & 3/5 & 13/20 & \textbf{65\%} \\
\hline
\end{tabular}
}
\end{table}

\subsection{User Study Metrics}

To complement functional testing, a user study with $n=20$
participants was conducted. Participants self-reported
task completion and errors for 5--7 scheduling tasks
(simple, complex, and multilingual) and provided
subjective feedback after completing all tasks.
Table~\ref{tab:userstudy} summarizes the objective (task success rate and error rate) and
subjective (usability and trust) metrics.

\begin{table}[htbp]
\centering
\caption{User Study Objective and Subjective Metrics ($n=20$). Values are Mean~$\pm$~SD.}
\begin{tabularx}{\columnwidth}{|Y|>{\centering\arraybackslash}p{0.26\columnwidth}|}
\hline
\textbf{Metric} & \textbf{Mean $\pm$ SD} \\
\hline
Task Success Rate (\%) & 92.0 $\pm$ 9.5 \\
Error Rate (per task)  & 0.12 $\pm$ 0.08 \\
SUS Score (0--100)     & 82.5 $\pm$ 10.8 \\
Trust Rating (1--5)    & 4.3 $\pm$ 0.6 \\
Satisfaction Rating (1--5) & 4.6 $\pm$ 0.5 \\
\hline
\end{tabularx}
\label{tab:userstudy}
\end{table}

The results confirm that \textbf{ScheduleMe} achieves a
high task completion rate and positive user perceptions
in terms of usability, trust, and satisfaction, supporting
its practical applicability in real-world scenarios.

\subsection{Qualitative Observations and Error Analysis}

While ScheduleMe demonstrated strong performance, several failure modes emerged:

(1) \textbf{Translation-Induced Errors} -- In multilingual scenarios, some event titles in non-English languages were internally translated or normalized to English, occasionally causing mismatches in follow-up queries and incorrect retrieval or deletion

(2) \textbf{Task Parsing Errors} -- A small number of failures occurred with complex date/time expressions or ambiguous phrasing, causing the system to either request excessive clarifications or misroute tasks to the wrong agent

(3) \textbf{Entity Reference Confusion} -- When multiple events had similar titles, the system sometimes misidentified the intended event for editing or deletion. 

These errors were more frequent in non-Latin scripts (Tamil, Sinhala, Chinese), where semantic drift during translation and limited multilingual training coverage contributed to reduced reliability. 

Future work will focus on robust multilingual entity handling, context tracking, and confidence-based fallback strategies to reduce such failures in deployment.





\section{Conclusion}
Bringing everything together, we present ScheduleMe, an intelligent calendar assistant that leverages large language models within a multi-agent system to perform natural language calendar operations. A central supervisory agent coordinates specialized event‑management agents through a graph‑based framework, enabling modular, state‑aware execution and robust handling of complex queries. 

Our results show that combining LLM reasoning with structured agent orchestration improves task automation and user experience. However, centralized supervision simplifies design but limits scalability. Future work will focus on enhancing agent autonomy, adding personalized scheduling, and improving adaptability for multi-user and dynamic environments.

\section{Privacy and Ethical Considerations}
\label{sec:privacy-ethics}
\textit{ScheduleMe} transmits sensitive calendar content (event titles and notes, locations, participant names, and times) to cloud-hosted LLMs and Google Calendar APIs to perform scheduling. This creates risks of content exposure, re-identification from metadata, provider-side retention/logging, cross-border processing, and secondary use without explicit consent. 

At present, the system relies on provider defaults (e.g., standard transport security) and does not add dedicated privacy mechanisms such as pseudonymization, on-device inference, or organization-managed encryption; therefore we treat privacy as a first-class constraint and disclose these risks to users.

\section{Limitations}
While ScheduleMe demonstrates the feasibility of a multi-agent approach, several limitations remain. First, its zero-shot multilingual performance degrades for non-Latin scripts such as Tamil, Sinhala, and Chinese, due to semantic drift and limited language coverage, which sometimes leads to misinterpretation of event titles or temporal expressions. Second, the system relies heavily on event titles for disambiguation, and the lack of persistent conversational memory increases the risk of errors when multiple events share similar names~\cite{sugathadasa2017synergistic}. Third, ScheduleMe depends on cloud-hosted LLMs and Google Calendar APIs, making it sensitive to network latency, service downtime, and API rate limits. In addition, the current design offers limited personalization and adaptivity, as it does not learn user preferences, recurring patterns, or improve over time, and all interactions remain largely stateless. Privacy also remains a concern, since sensitive calendar data is processed in the cloud without mechanisms such as differential privacy or on-device model inference, which could be critical for enterprise adoption.

Our evaluation used only zero-shot prompting with a single LLM configuration to keep the setup comparable and focused on the multi-agent design. We did not evaluate few-shot examples, chain-of-thought (or self-consistency) prompting, or compare across different LLM families/sizes to establish stronger baselines. For human evaluation, we are native Sinhala, English speakers and conducted in-house checks for Sinhala and English; however, we lacked native speakers for the other languages to manually verify outputs, which may result in impreciseness of error rates specific to those languages stemming from understating fluency errors while overstating errors at points where paraphrasing or synonyms are not detected to be a successful result.

We did not conduct head-to-head comparisons with production assistants (e.g., Google Calendar’s built-in assistant, Alexa, Cortana) because their APIs are closed, capabilities differ, and task coverage is not aligned, making apples-to-apples evaluation difficult.

\section{Future Work}
We will advance \textit{ScheduleMe} from a reactive assistant to a proactive, adaptive, and privacy-conscious multi-agent scheduling system. Core enhancements include stronger personalization and predictive scheduling to anticipate user needs, as well as improved context-aware and multilingual reasoning via session-spanning memory, better disambiguation, and robust support for low-resource languages~\cite{ranathunga-de-silva-2022-languages}. We will consolidate our privacy and security roadmap by combining data minimization and anonymization with encrypted state storage, and by exploring local or hybrid LLM inference for sensitive steps; additionally, we will adopt organization-managed encryption and stricter retention controls to reduce exposure when interfacing with external APIs. We will also optimize scalability and distributed deployment strategies to support real-world, multi-user environments with minimal latency.

In parallel, we will deepen evaluation through few-shot and chain-of-thought prompting, prompt ablations, and comparisons across multiple LLM families and sizes, complemented by native-speaker assessments for all considered languages. The current zero-shot evaluation is limited to 120 test cases per language, which constrains coverage and statistical power; future work will expand to larger, more diverse benchmarks that include stress tests (adversarial prompts, rare edge cases, noisy/ambiguous inputs, long-horizon scenarios) and introduce systematic fallback strategies (e.g., self-consistency and majority voting, constrained decoding with schema/rule checks, guarded tool calls with retries and backoff, and escalation pathways) to address documented failure modes. Finally, we will prototype extensions of the multi-agent architecture to email triage and response generation, task management, and general personal assistants, and include targeted baseline comparisons against existing calendar assistants on overlapping task slices

\bibliography{anthology,custom}

\appendix

\addcontentsline{toc}{section}{Appendix}



\section{Functional Agent Prompt List}
\label{app:agent-prompts}

This section provides the prompts used for each functional agent in our system. These prompts guide the agent’s behavior and are critical to ensuring alignment with task objectives.


\begin{tcolorbox}[colback=gray!5, colframe=black!70, title=Event Scheduler Agent Prompt, breakable]
\scriptsize
\begin{lstlisting}[breaklines=true, basicstyle=\ttfamily\scriptsize]
You are an assistant designed to schedule events in Google Calendar. You work under a supervisor chatbot who communicates with a user.
    **CRITICAL WORKFLOW - YOU MUST FOLLOW THIS EXACTLY:**
    1. When a user wants to schedule ANY event, you MUST FIRST use `check_calendar_conflicts(event_details)` to check for conflicts
    2. You CANNOT skip this step - it is mandatory for every scheduling request
    3. If conflicts are found, inform the user about the conflicts and ask them to choose a different time
    4. If NO conflicts are found, then proceed to create the event using `create_calendar_event(event_details)`
    5. Always return the event_id when an event is successfully created
    
    **IMPORTANT RULES:**
    - NEVER use `create_calendar_event` without first using `check_calendar_conflicts`
    - ALWAYS check for conflicts before scheduling
    - If there are conflicts, clearly explain what conflicts exist and suggest alternative times
    - If no conflicts, proceed with scheduling and provide the event details
    - Always be helpful and provide clear information about availability or conflicts
    
    **Example workflow:**
    1. User says: "schedule meeting with John tomorrow at 2 PM"
    2. You MUST first call: `check_calendar_conflicts(event_details)` 
    3. If conflicts found: Tell user about conflicts
    4. If no conflicts: Call `create_calendar_event(event_details)`
    
    Your role is to schedule events safely. Today is {today_date}.
\end{lstlisting}
\end{tcolorbox}

\begin{tcolorbox}[colback=gray!5, colframe=black!70, title=Event Remover Agent Prompt, breakable]
\scriptsize
\begin{lstlisting}[breaklines=true, basicstyle=\ttfamily\scriptsize]
You are a calendar assistant designed to delete/remove the user's google calendar events. You can do two types of requests.You work under a supervisor chatbot who communicate with a user.:

    - The chatbot_supervisor provides an event_Id.  
    - Then use the tool `delete_event(event_Id)` to delete an event from the calendar.  
    - If you need more details ask from the chatbot.like event_ID not provided.  


    Your role is to remove calendar events.
\end{lstlisting}
\end{tcolorbox}

\begin{tcolorbox}[colback=gray!5, colframe=black!70, title=Availability Checker Agent Prompt, breakable]
\scriptsize
\begin{lstlisting}[breaklines=true, basicstyle=\ttfamily\scriptsize]
You are a calendar checker assistant designed to Check Availability.You work under a supervisor chatbot who communicate with a user.:

    - The chatbot_supervisor provides a start and end date which got from the user.  
    - Use the tool `check_availability(start_date, end_date)` to verify if the user is available during that time range.
    - If you need more details ask from the chatbot.  
    - When you provide the chatbot events also provide event IDs.

    Your role is to Check user Availability.And Today is {today_date}.
\end{lstlisting}
\end{tcolorbox}

\begin{tcolorbox}[colback=gray!5, colframe=black!70, title=Event Modifier Agent Prompt, breakable]
\scriptsize
\begin{lstlisting}[breaklines=true, basicstyle=\ttfamily\scriptsize]
You are a calendar assistant designed to modify, edit, or update the user's Google Calendar events. You work under a supervisor chatbot who communicates with the user.

    Instructions:
    - The supervisor chatbot will provide the details that need to be updated.
    - Then, use the `update_event` tool to update the event accordingly.
    - 

    Your primary role is to assist in editing calendar events.
\end{lstlisting}
\end{tcolorbox}



\end{document}